\documentclass[Times2COL]{WileyNJDv5} 

\articletype{Original Research Paper}%

\received{ }
\revised{ }
\accepted{ }
\volume{pages}
\copyyear{2026}
\startpage{1}

\usepackage{tikz}
\usetikzlibrary{arrows.meta, positioning, fit, calc}
\usepackage{balance}

\begin{document}

\title{Sensor-Driven Mission Synthesis for UAV/UGV Swarms: \\ A TB-CSPN Coordination Architecture with \\ Hardware-Enforced Safety}

\author[1]{Uwe M. Borghoff}

\author[2]{Paolo Bottoni}

\author[3]{Remo Pareschi}

\authormark{BORGHOFF, BOTTONI \& PARESCHI}
\titlemark{SENSOR-DRIVEN MISSION SYNTHESIS FOR UAV/UGV SWARMS}

\address[1]{\orgdiv{Institute for Software Technology}, \orgname{University of the Bundeswehr Munich}, \orgaddress{\state{85579 Neubiberg}, \country{Germany}}}

\address[2]{\orgdiv{Department of Computer Science}, \orgname{Sapienza University of Rome}, \orgaddress{\state{00161 Rome}, \country{Italy}}}

\address[3]{\orgdiv{SofTware And Knowledge Engineering Lab}, \orgname{STAKE lab}, 
\orgaddress{\state{86100 Campobasso}, \country{Italy}}}

\corres{Corresponding author Uwe M. Borghoff \email{uwe.borghoff@unibw.de}}

\abstract[Abstract]{This paper presents a coordination architecture for heterogeneous UAV/UGV swarms that synthesises mission actions from uncertain, multi-modal sensor evidence while preserving hardware-enforced safety at the actuation boundary. 
The approach combines radar, RF, acoustic, and visual observations with 
Topic-Based Communication Space Petri Net (TB-CSPN) orchestration to support incremental mission formation under partial and evolving information. 
Consultant agents transform sensor outputs into temporally bounded semantic tokens, while supervisor agents provide authorisation and policy-governed release of mission transitions. 
This separation between interpretation, coordination, and execution yields auditable decision paths, constrains non-determinism within the coordination layer through guards and synchronisation, and enables bounded-time integration of heterogeneous evidence. 
To improve resilience in contested environments, including cyber compromise, spoofing, jamming, and communication loss, the digital coordination layer is complemented by independent analogue safety envelopes that clamp or veto unsafe actuator commands issued to individual vehicles. 
A coastal-surveillance case study illustrates how the proposed architecture enables dependable, governed, and physically safe swarm coordination under operational uncertainty.}


\maketitle

\renewcommand\thefootnote{}
\footnotetext{\textbf{Abbreviations:} TB-CSPN, Topic-Based Communication Space Petri Net; UAVs/UGVs, unmanned aerial and ground vehicles.}

\renewcommand\thefootnote{\fnsymbol{footnote}}
\setcounter{footnote}{1}

\section{Introduction}\label{sec:intro}
Swarms of unmanned aerial and ground vehicles (UAVs/UGVs) are attracting interest due to their offering of ``enhanced intelligence, improved coordination, increased flexibility, survivability, and reconfigurability'' \cite{iotj/JavedHAAASG24}.
In this line, we propose the Guarded Swarms \cite{techdefense/BorghoffBP25} hybrid architecture for coordinating UAV and UGV swarms, combining digital autonomy with analogue (hardware) safety enforcement at the actuator level, so as to keep high-level decision-making and mission coordination in software, where AI-based perception modules and formal workflow models can adapt to changing operational conditions, while moving the most safety-critical “do-not-ever-do-this” constraints into hardwired analogue circuits placed between the vehicle’s flight or drive controller and its motors. 

This architecture creates a two-layer assurance model: flexible, semantically informed coordination above, and irreducible, attack-resistant safety guarantees below. 
The digital layer interprets sensor signals, forms mission plans, and coordinates swarm actions, while the analogue one prevents unsafe commands from reaching the actuators, even in the presence of software faults, adversarial interference, or communication failures.

\begin{figure*}[ht]
\centering
\includegraphics[width=\textwidth]{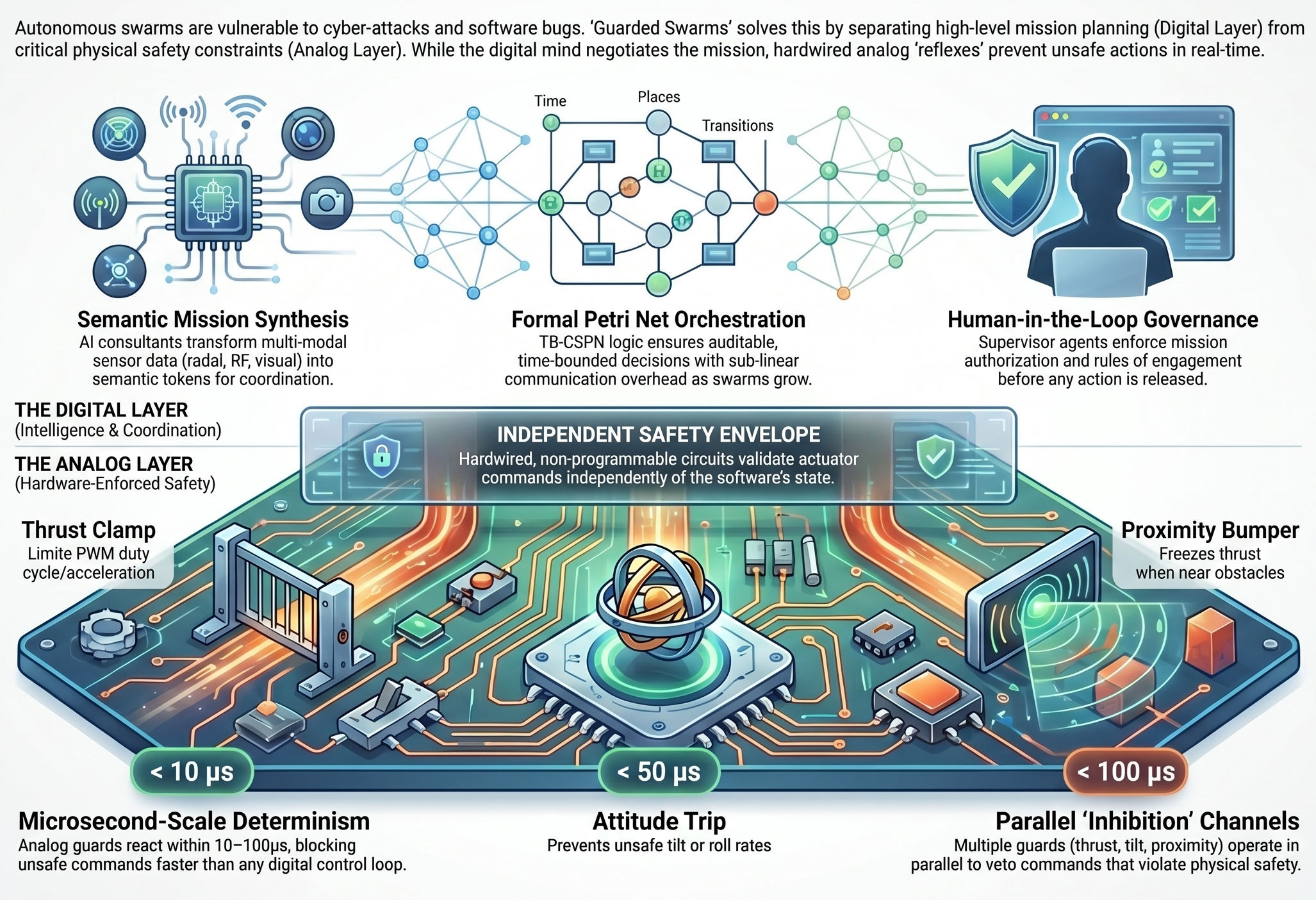}
\caption{Conceptual overview of the Guarded Swarms architecture. 
The digital layer transforms multi-modal sensor inputs into semantic tokens and coordinates mission formation through TB-CSPN, exhibiting constrained non-determinism: multiple admissible coordination paths may exist, but are strictly bounded by token availability, guard conditions, and temporal constraints. 
Human-in-the-loop governance enforces authorisation policies before mission release. 
The analogue safety envelope provides independent, hardware-enforced veto mechanisms that ensure deterministic, physically safe actuation.}
\label{fig:conceptual}
\end{figure*}

To provide a high-level conceptual view of the proposed architecture, Figure~\ref{fig:conceptual} illustrates the separation between semantic coordination in the digital layer and hardware-enforced safety in the analogue layer, together with the role of human-in-the-loop governance and bounded coordination dynamics.

Building on the TB-CSPN framework proposed in previous work, we separate semantic interpretation from coordination execution. 
In TB-CSPN, large language models (LLMs) or specialised AI modules extract topics and intents, while Coloured Petri Nets / Workflow Nets provide formally analysable coordination logic with constrained non-determinism:  multiple transitions may be enabled, but the set of admissible evolutions is strictly bounded by token availability, guard conditions, and temporal constraints \cite{jcsc/Aalst98}. 
TB-CSPN also claims scalability advantages (sub-linear coordination overhead as swarm size increases). 
We extend this lineage by introducing an ``executor'' role, e.g., physical robots or vehicles, responsible for translating the commands issued by the net, and reading from specific output places.
Such executors operate inside analogue safety envelopes, thereby extending agentic coordination into the physical domain, under the architectural boundaries of Table~\ref{tab:boundaries}.

\begin{table}[ht]
\caption{Architectural boundaries and their primary guarantees.}
\begin{tabular}{ll}
\toprule
\textbf{Boundary} & \textbf{Primary Guarantee} \\
\midrule
Human--AI & Accountability and governance \\
Role separation & Clear authority and delegation \\
Semantic--coordination & Scalable, verifiable orchestration \\
Digital--physical & Deterministic safety \\
\bottomrule
\end{tabular}
\label{tab:boundaries}
\end{table}

The motivation is operational: in contested environments (electronic warfare, jamming, spoofing, cyber compromise, comms denial), purely digital autonomy is fragile because software and networks are part of the attack surface. Guarded Swarms reduce that exposure by shifting critical guardrails into hardware that operates independently of digital computation and communications. Digital planners (centralised or federated) may propose trajectories and manoeuvres, but analogue boards validate, clamp, veto, or override low-level motor commands before they reach electronic speed controllers. Once a command is admitted through the envelope, execution becomes deterministic, but unsafe commands cannot propagate—even if the software stack is degraded or maliciously influenced.

Concretely, each vehicle includes an Analogue Safety Board implementing various independent enforcement channels. Examples include: thrust clamps (window comparators limiting PWM duty cycle and ramp rates), attitude/rate trips (analogue gyro/tilt thresholds triggering clamp or cut), proximity ``bumpers'' (IR/ultrasonic comparators biasing or freezing thrust near obstacles), power/thermal protection (current/temperature comparators forcing shutdown), RF guard tone mechanisms (loss of tone or kill-code triggers land/kill), and a hard E-stop using opto-isolated latches and manual switches. 
Importantly, these are not a serial pipeline; they operate in parallel with an OR-style inhibition logic: if any guard triggers, it can unilaterally intervene. 
If none triggers, the PWM passes unchanged. 

\begin{figure}[ht]
\centering\includegraphics[width=\columnwidth]{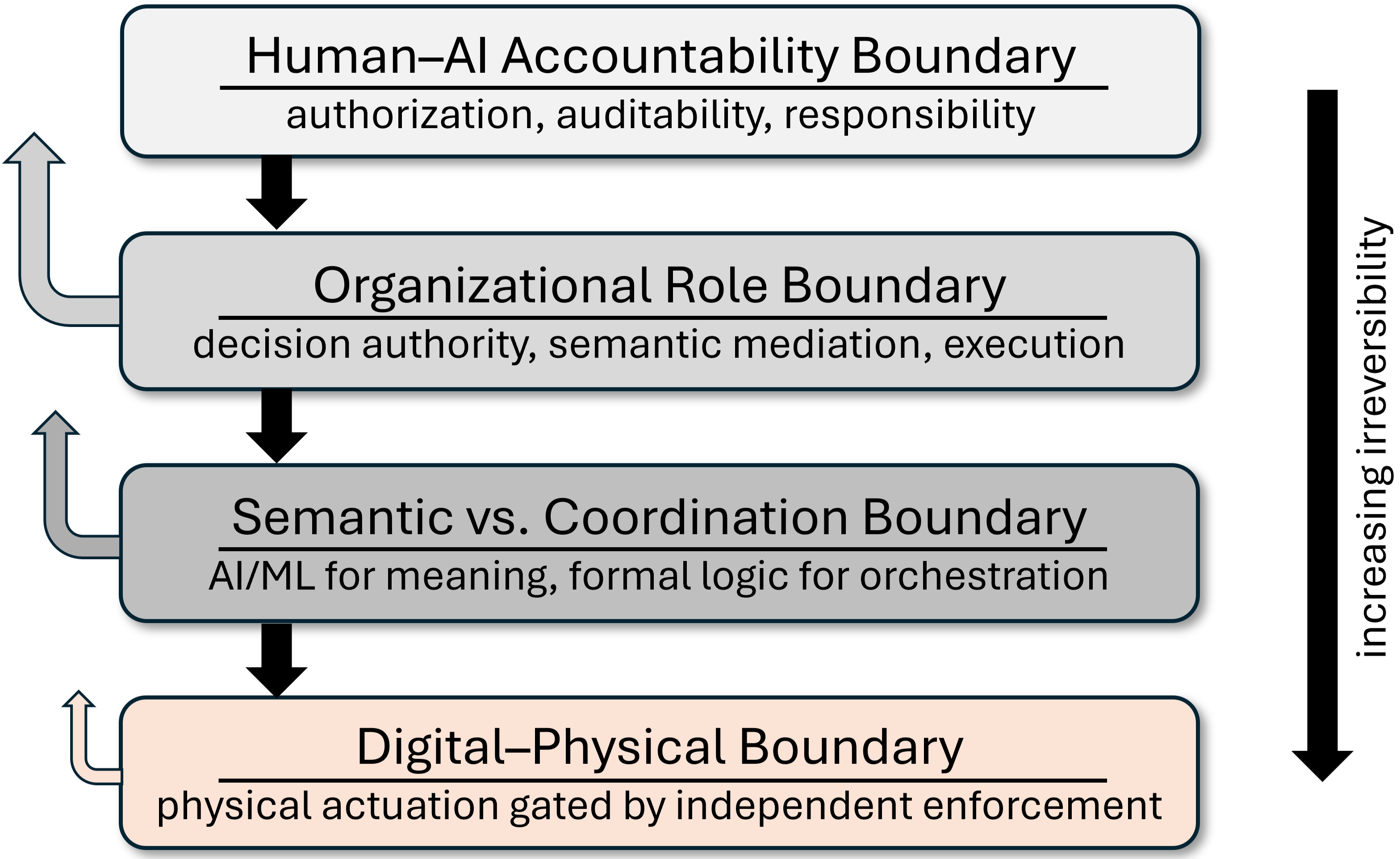}
\caption{Architectural boundaries in autonomous AI systems.
As decisions propagate downward, their reversibility decreases and the cost of error increases, motivating progressively stronger guarantees.}
\label{fig:boundaries}
\end{figure}

We frame this as analogous to contract-based design: the digital layer can vary, but the analogue one enforces a fixed safety ``contract'' \cite{FUTUREINTERNET/BorghoffBP26}.
As decisions move downward in the architecture (see Figure~\ref{fig:boundaries}), reversibility decreases. 
In other words, the ability of feedback to reach and influence a higher level decreases, thereby preventing situations that are not ``contracted''.

We build on this foundation, shifting the focus from formal correctness to real-world applicability, systems engineering, and empirical demonstration. 
Hardware-level security primitives, e.g., physical unclonable functions (PUFs)~\cite{gao2020physical}, trusted platform modules (TPMs)~\cite{platform2020tpm}, hardware security modules (HSMs) \cite{ictc/KimK20d}, trusted execution environments (TEEs) \cite{csur/PintoS19}, secure-boot roots of trust \cite{sp/ParnoMP10}, remote (hardware) attestation mechanisms \cite{compsec/KuangFSYG22}, and anti-tamper, side-channel, or fault-injection countermeasures \cite{crypto/Schneider0G16}, are beyond the scope of this paper.

The Guarded Swarms architecture was first introduced at IEEE TechDefense 2025~\cite{techdefense/BorghoffBP25}. Two concurrent developments followed, both rooted in that conference article. The companion paper~\cite{FUTUREINTERNET/BorghoffBP26}, in \emph{Future Internet}, focuses on the architecture's lower boundary: the interaction between digital coordination and the analogue safety layer at the actuation side. The present paper develops the upper boundary: the acquisition, interpretation, and coordination of multi-modal sensor evidence on the input side. The two papers are therefore complementary by design rather than overlapping, and together they trace the full path of evidence and authority through the architecture, from heterogeneous sensing through governed coordination to physically safe actuation. Within this division of labour, the present paper contributes (i) an extension of the TB-CSPN token model with the temporal and provenance metadata required for fusing heterogeneous sensor evidence, (ii) time-windowed synchronisation as the mechanism through which asynchronous detections become coordination-ready, (iii) a coastal-surveillance case study illustrating incremental mission formation from radar, RF, acoustic, and electro-optical inputs, and (iv) a discussion of concrete protocol integration with prevailing middleware stacks (\textsc{ROS~2}, \textsc{DDS}, \textsc{MAVLink}, \textsc{ASTERIX}).

The remainder of this paper is organised as follows. Section~\ref{sec:motivation} motivates the proposed approach, and Section~\ref{sec:contributions} summarises the main contributions and the underlying architecture. Related work is reviewed in Section~\ref{sec:relwork}. Section~\ref{sec:case_study} illustrates the approach through a case study, including counterfactual trajectories that show how the architecture's structural guarantees operate. Section~\ref{sec:discussion} discusses concrete protocol integration with prevailing middleware stacks, and Section~\ref{sec:discussion-proper} reflects on the architecture's commitments and trade-offs. Finally, Section~\ref{sec:conclusion} concludes the paper and outlines directions for future work.

\section{Motivation: Sensor-Driven Coordination Under Uncertainty}\label{sec:motivation}
Autonomous UAV/UGV systems operating in contested or complex environments must make decisions based on incomplete, heterogeneous, and continuously evolving evidence.
More generally, drone swarms have been characterised as fundamentally information-dependent systems, since their coordination, adaptability, and operational effectiveness rely on robust communication and the efficient processing of distributed information~\cite{kallenborn2022infoswarms}.

In many operational settings, signals originate from multiple sensing modalities, such as radar tracks, RF emissions, acoustic signatures and visual confirmations. Each of these modalities is characterised by a distinct profile of latency, reliability and ambiguity.

This poses a critical engineering challenge. 
Decisions must often be made under tight time constraints, with available evidence often partial or even contradictory. 
Premature commitment to a single interpretation can lead to incorrect mission responses, while excess hesitation may prevent timely action.

Recent multi-sensor UAV detection work has likewise addressed conflicts among heterogeneous detector outputs through probabilistic fusion mechanisms to improve identification \cite{apin/SaadaouiCD23}, while related IET Radar, Sonar \& Navigation research has demonstrated Bayesian information fusion for maritime situational awareness using heterogeneous radar sensors \cite{gaglione2020bayesian}.
 
The goal is not just to compute a single ``best explanation'', but to maintain a \emph{structured operational state} that
\begin{itemize}
\item preserves competing interpretations of sensor evidence,
\item makes compatibility and tension among signals explicit,
\item supports incremental refinement as new observations arrive, and
\item enables safe, authorised actions under bounded uncertainty.
\end{itemize}

The architecture proposed in this paper addresses this challenge by combining multi-source signal interpretation with structured coordination semantics. Rather than relying on a monolithic planner, the system transforms heterogeneous sensor detections into \emph{semantic tokens} and synchronises them within a TB-CSPN coordination space. Mission-level decisions emerge incrementally as evidence accumulates and authorisation constraints are satisfied.

This approach is particularly suited to environments in which sensing, decision-making, and actuation must remain auditable and time-bounded. By separating signal interpretation from coordination logic and from actuation enforcement, the architecture supports dependable operation even when evidence remains incomplete or evolving.

\section{Main Contributions}\label{sec:contributions}
This paper makes four contributions relevant to engineering audiences concerned with dependable autonomy, multi-source sensing, and coordinated robotic systems.
\newline

\noindent\textbf{C1: Sensor-driven mission synthesis through TB-CSPN coordination.}
TB-CSPN provides coordination semantics with \emph{constrained non-determinism}: while multiple transitions may be enabled, the set of admissible evolutions is strictly bounded by token availability, guard conditions, and temporal constraints. This includes both structurally bounded alternatives (in which all possible outcomes are known in advance) and operationally indifferent choices (in which alternative outcomes are equivalent with respect to mission objectives). 
These semantics are combined with time-windowed synchronisation and explicit mission formation transitions that convert evolving evidence into coordinated swarm actions.
\newline

\noindent\textbf{C2: Temporal coherence as a coordination contract for sensor fusion.}
Multi-source sensor fusion is fundamentally a problem of \emph{temporal alignment}: detections from radar, RF, acoustic, and electro-optical modalities arrive asynchronously, with varying latencies, validity intervals, and confidence levels. We define digital coordination contracts that enforce temporal coherence across heterogeneous evidence: each token carries an explicit time-to-live and provenance, and mission-formation transitions fire only when the participating tokens fall within a bounded synchronisation window. This ensures that mission decisions reflect a temporally consistent evidence set, rather than the opportunistic arrival order of sensor messages. The contract mechanism is general, but its motivation here is specific to multi-modal sensor coordination, where premature fusion across stale evidence is a documented failure mode~\cite{apin/SaadaouiCD23, gaglione2020bayesian}.
\newline

\noindent\textbf{C3: Structured governance of mission formation.}
The architecture incorporates supervisor agents that enforce operational policies, including rules of engagement, mission constraints, and resource availability. TB-CSPN provides explicit authorisation transitions that capture human or policy oversight as a formal coordination mechanism rather than an informal external process.
\newline

\noindent\textbf{C4: Application to multi-modal sensing and guarded swarm autonomy.}
We demonstrate the architectural pattern through a representative case study involving heterogeneous radar, RF, and acoustic signals used to coordinate UAV/UGV swarm responses. The example illustrates how incremental signal interpretation can lead to mission formation while remaining consistent with the Guarded Swarms principle of hardware-level actuation safety.

\subsection{Architectural Overview}
\begin{figure*}[t]
\centering\includegraphics[width=\textwidth]{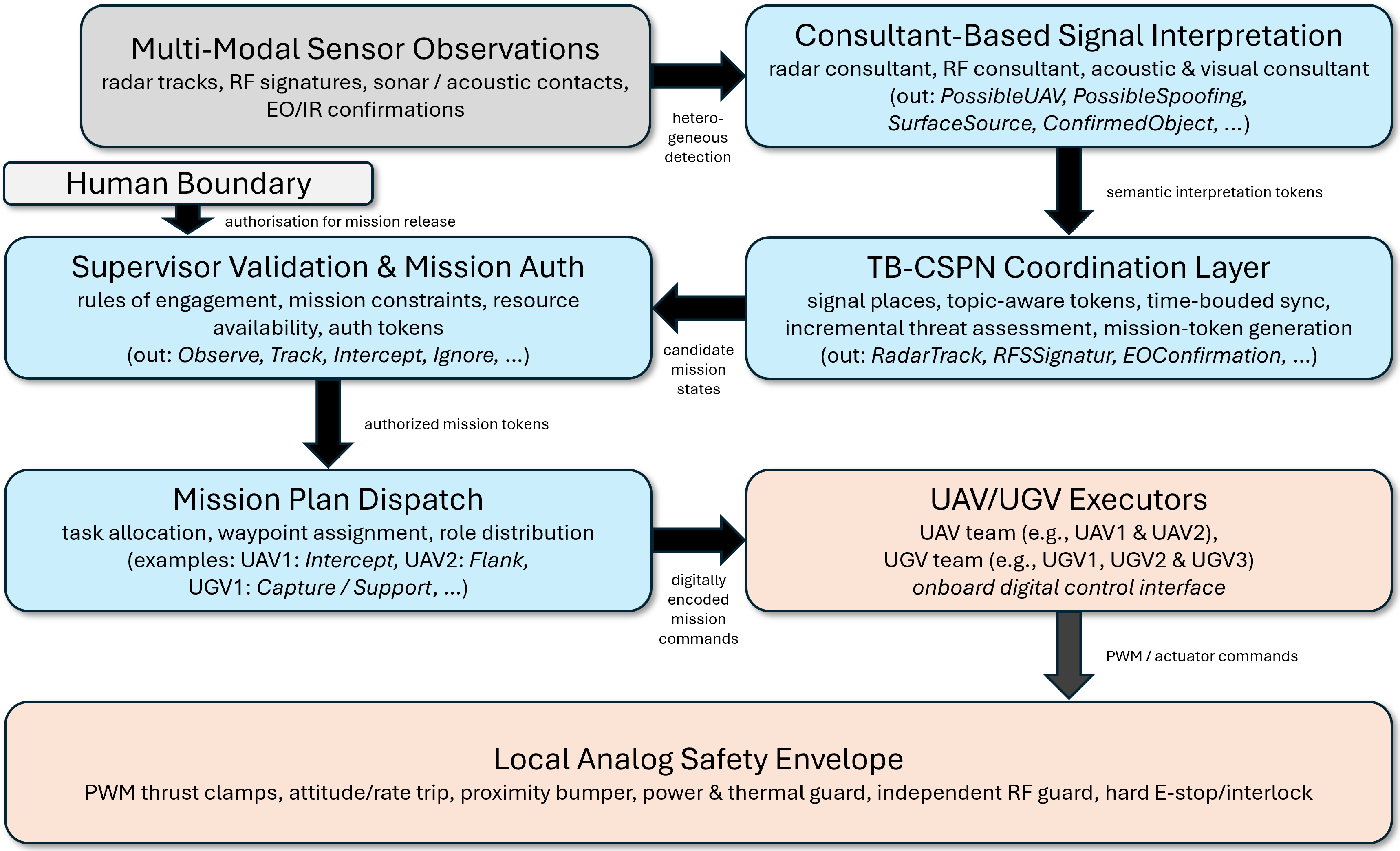}
\caption{Overall architecture for sensor-driven mission synthesis in guarded UAV/UGV swarms.}
\label{fig:architecture}
\end{figure*}

Figure~\ref{fig:architecture} shows the layered structure of the Guarded Swarms architecture. 
It comprises four functional stages that transform raw multimodal sensor observations into safe swarm actions.
\newline

\noindent\textit{Evidence producers.}
Sensor systems generate observations such as radar detections, RF signatures, acoustic contacts, and visual confirmations \cite{ijcip/SemenyukKLAKC25}. 
Each observation carries provenance, timestamp, and confidence metadata.
For an implementation of a multiple‐input multiple‐output (MIMO) radar system developed for small drones, see \cite{Yang2021}. For a review of general radar classification, see \cite{Patel2018}.
\newline

\noindent\textit{Consultant agents for signal interpretation.}
Specialised agents interpret raw observations to produce semantic tokens describing possible objects or events (e.g., potential UAV presence, surface-source detection, or signal interference). 
These agents may incorporate statistical models, signal processing modules, or learned classifiers.
\newline

\noindent\textit{TB-CSPN coordination layer.}
Interpreted signals are represented as typed tokens that enter a TB-CSPN coordination space. The Petri-net orchestration layer synchronises tokens across bounded time windows and generates mission-level tokens through explicit transitions. This coordination space serves as a structured, auditable backbone with constrained non-determinism that governs the evolution of evidence into operational decisions.
\newline

\noindent\textit{Supervisor authorisation and mission formation.}
Supervisor agents enforce operational policies, including authorisation requirements, mission constraints, and swarm resource allocation. 
This explicit authorisation step is consistent with broader arguments for maintaining meaningful human control over autonomous systems in high-stakes settings \cite{firai/SioH18}.
This emphasis on explicit human authorisation is consistent with recent work showing that AI can improve analysis quality in high-stakes military settings while still requiring human oversight, particularly under ambiguous and contradictory information~\cite{NitzlCKB25}.
Only when both evidence and authorisation tokens are present can mission transitions fire.
\newline

\noindent\textit{Execution and safety enforcement.}
Authorised missions are dispatched digitally to worker agents associated with the UAV and UGV executor platforms.
Worker agents consume mission tokens produced by the coordination layer and translate them into vehicle-specific control instructions for the corresponding executors.
Hence, they perform deterministic translation and execution, in contrast to the constrained non-determinism of the coordination layer, and are barred from mission-level reasoning or hypothesis generation. 
\newline

At the actuation boundary, each executor is protected by an independent analogue safety envelope that can clamp or veto unsafe commands, ensuring that mission-level reasoning and physical safety remain complementary but independent.

Together, these roles implement a layered coordination structure in which interpretation, governance, and physical execution remain explicitly separated.

In principle, each place can be associated with a user interface element that allows humans to inspect the tokens it contains over time.
Similarly, suitable views of transitions allow inspection and (traced) editing of their firing conditions and of the expressions (possibly prompts to generative agents) used to compute the values of the produced tokens.

\subsection{Token-Based Data Model}
Information exchange within the coordination layer occurs via typed tokens inspired by coloured Petri net semantics, timestamped with the time $t$ of their generation, and augmented with a temporal validity value (``Time To Live'', $ttl$).
\newline

\noindent\textit{Observation tokens.}
Sensor observations are represented as tokens carrying provenance and validity metadata:
\[
\mathsf{ObsToken} = \langle id, value, source, confidence, t, ttl \rangle
\]

These tokens reify, and bring inside the digital control subsystem, the signals produced by the sensors, e.g., radar tracks or RF detections. 
\newline

\noindent\textit{Interpretation tokens.}
Consultant agents generate interpretation tokens describing semantic interpretations, expressed as labels of the information carried by the observation tokens:
\[
\mathsf{InterpToken} = \langle label, origin, confidence, t, ttl \rangle
\]

Examples of labels include \textit{PossibleUAV}, \textit{PossibleSpoofing}, or \textit{SurfaceSource}.
\newline

\noindent\textit{Mission tokens.}
When sufficient evidence and authorisation are present, mission tokens are generated:
\[
\mathsf{MissionToken} =
\langle mission\_id, type, target, justification, t, ttl \rangle
\]

These tokens trigger coordinated swarm actions.

\subsection{Workflow Semantics: TB-CSPN Coordination}

TB-CSPN provides the orchestration backbone, leading tokens to evolve into actions. 
Three classes of constraints are enforced.
\newline

\noindent\textit{Time-windowed synchronisation.}
Sensor evidence from different sources can be fused within bounded temporal windows to ensure that coordinated decisions are based on temporally coherent observations.
This means that a token with $t=time_0$ can contribute to the analysis until time $time_0+ttl$, possibly combined with other tokens for which $abs((t^\prime_0+ttl^\prime)-(t_0+ttl))\le{ths}$, for some threshold $ths$ established depending on the type of decision to be made.
\newline

\noindent\textit{Guarded mission formation.}
Designated transitions fire only when the required combination of interpretation and authorisation tokens is present (within the synchronisation window). 
This ensures that mission generation follows explicit coordination rules under constrained non-determinism.
\newline

\noindent\textit{Governed action release.}
Authorisation tokens provided by supervisor agents are required for mission execution. This formal coordination mechanism captures operational governance, such as rules of engagement and resource availability.

\subsection{Application Pattern: Multi-Modal Threat Detection}
The architecture applies naturally to multi-modal threat detection scenarios in which heterogeneous sensing signals must be interpreted incrementally. Radar tracks, RF emissions, and acoustic contacts may initially support multiple interpretations. As additional signals arrive, the coordination space progressively refines the interpretation until a mission decision (e.g., \textit{act} or \textit{abort}) becomes justified.
In this way, the system maintains a structured interpretation landscape while still enabling timely responses. 
The case study presented in Section~\ref{sec:case_study} demonstrates this process in a representative coastal surveillance scenario involving coordinated UAV/UGV swarm actions.

\subsection{Addressing Practical Coordination Constraints}
The architecture proposed in this paper is designed to operate under realistic cyber-physical constraints, including communication latency, partial observability, and the need to resort to hardware-level security measures. 
There are several design principles that help to ensure the robustness and trustworthiness of the coordination model under such conditions.
For a review of similar hardware-rooted trust mechanisms, see~\cite{fi/AhmadSA25}.
\newline

\noindent\textit{Local Security Overrides and Collective Behavior.}
In the supervised swarm paradigm, each vehicle is protected by an analogue security envelope that can block unsafe actuator commands. Although these local security measures may occasionally override the digitally generated swarm directives, the coordination layer is designed to tolerate limited deviations from the nominal plan. Swarm behaviors are therefore treated as \textit{robust collective strategies} rather than rigid synchronised movements. In practice, this means that security overrides can temporarily interrupt local trajectories without compromising the overall mission objective.
\newline

\noindent\textit{Communication latency and distributed operation.}
Swarm elements can operate under intermittent or delayed communication conditions. The TB-CSPN coordination layer, therefore, relies on limited synchronisation windows rather than continuous message exchange. Vehicles execute local control policies between coordination events, keeping the swarm partially autonomous even under degraded connectivity. Mission updates are enacted whenever new tokens become available, and coordination transitions are consequently triggered, rather than relying upon constant global synchronisation.
\newline

\noindent\textit{Two-phase mission formation.}
Mission generation follows a two-phase process. 
First, interpreted sensor signals enable the creation of candidate mission-related tokens within the coordination space. Second, supervisor agents provide authorisation tokens that enable mission release transitions. This separation between mission proposal and mission authorisation aligns with two-phase commit mechanisms, constraining rules of engagement, resource availability, and security policies to be satisfied before execution.
\newline

\noindent\textit{Hierarchical Control Structure.}
The architecture naturally supports a two-level control model. The coordination layer generates mission-level directives such as objective assignments, patrol regions, or formation configurations. Individual UAV and UGV platforms then execute these directives through local autonomy modules responsible for trajectory generation and stabilisation. If communication with the coordination layer is interrupted, the vehicles revert to safe local behaviors, which are loosened only upon receipt of updated mission tokens.
\newline

\noindent\textit{Agent Roles and Execution Semantics.}
Within the TB-CSPN model, different classes of agents interact with the coordination space in distinct ways. Consultant agents interpret sensor observations and generate semantic tokens. Supervisor agents enforce policy constraints and enable authorisation transitions. Worker agents do not control coordination transitions; instead, they use mission tokens produced by the orchestration layer and translate them into vehicle-specific control commands. 
This separation cuts clear boundaries between reasoning, governance, and physical execution.
\newline

\noindent\textit{Temporal Consistency of Evidence.}
Tokens entering the coordination space contain timestamps and validity intervals. 
TB-CSPN transitions can therefore enforce time-window constraints, ensuring that coordinated decisions are based on temporally consistent observation sets. 
This mechanism prevents outdated sensor information from influencing mission formation while preserving the ability to verify decision history.
\newline

Together, these design principles allow the coordination architecture to remain effective in environments characterised by asynchronous sensing, partial communication, and stringent security requirements.

\section{Related Work}\label{sec:relwork}
The present study draws on ideas from our earlier work on trusted multi-agent coordination, presented at IEEE TechDefense~2025~\cite{techdefense/BorghoffBP25}, in which digital orchestration was combined with hardware-level safety enforcement. That effort followed the applied investigation in~\cite{carovilla2023integrating}, which employed blockchain-backed monitoring for unmanned systems. While effective for accountability, the centralised design highlighted bottlenecks and single-point vulnerabilities, motivating a shift toward distributed architectures. As an example, Zhou \textit{et al.}~\cite{Zhou2024} employ distributed block consensus as a key component of a blockchain-based adaptive networking mechanism for UAVs.

We gave a broader theoretical formulation of distributed coordination in~\cite{FRONTIERS/BorghoffBP25}, integrating concepts from multi-agent~\cite{Wooldridge2002} and Centaurian systems~\cite{centaurian24} within a communication space framework. 
This line of work led to the TB--CSPN architecture~\cite{DISCOVER/BorghoffBP25}, which separates semantic reasoning from coordination logic, with constrained non-determinism implemented via coloured Petri nets~\cite{Jensen2009}. 
The architecture supports both autonomous and human-supervised operation while ensuring key correctness properties such as liveness and bounded execution~\cite{FUTUREINTERNET/BorghoffBP25}.

A notable feature of TB--CSPN is its \emph{sublinear} coordination overhead~\cite{FUTUREINTERNET/BorghoffBP25}, which has been validated analytically and through open-source implementations (\url{https://github.com/Aribertus/tb-cspn-poc}). Hierarchical token routing and localised concurrency windows reduce communication load as swarm size increases. These scalability characteristics carry directly into the Guarded Swarms framework, where digital coordination remains lightweight, and the analogue enforcement layer introduces only constant-time actuator checks.

The Guarded Swarms architecture~\cite{FUTUREINTERNET/BorghoffBP26} introduces a hybrid architecture for resilient UAV and UGV coordination combining digital planning with independent analogue safety enforcement. 
High-level directives from supervisors and AI modules are processed by a scalable token-based coordination model, while simple, fast analogue circuits validate or veto motor commands. 
This separation ensures flexible orchestration and deterministic actuation at the physical layer, even under cyber compromise or communication loss, as illustrated via a two-robot UAV and UGV case study.
Lassfolk \textit{et al.}~\cite{Lassfolk2026} present a conceptual certificate-based two-layer trust model together with a drone-swarm onboarding procedure, motivated by a dynamic battlefield scenario.

Recent defence-oriented systems illustrate the growing convergence of software-defined mission logic, autonomous coordination, and embedded safety enforcement. 
For example, Helsing’s \textsc{HX-2} is presented as an AI-enhanced software-defined strike UAV with onboard avionics that support safety-relevant constraints such as geofencing, collision avoidance, and flight-envelope limitations (\url{https://helsing.ai/altra}).

Related programmatic efforts include \textsc{DARPA}'s \textsc{OFFSET} initiative, which explored the combination of centralised mission logic with distributed swarm execution (\url{https://www.darpa.mil/research/programs/offensive-swarm-enabled-tactics}), and the UK \textsc{LANCA} (``Loyal Wingman'') programme, where centrally issued manoeuvre directives are paired with autonomous onboard constraint enforcement (\url{https://thedefensepost.com/2022/11/03/uk-launches-combat-drone-project/}). 

On the ground-systems side, \textsc{ARX}'s \textsc{Mithra OS} exemplifies the digital transformation of legacy UGV fleets into interoperable and coordinated platforms for contested operational environments (\url{https://www.arx-robotics.com/mithra-os}). These examples are consistent with the broader architectural trend of separating mission-level coordination from low-level platform safeguards, although they generally do not provide the explicit token-based orchestration and formal coordination semantics developed in the present work.

Survey work has outlined broader adoption of AI and Deep Reinforcement Learning (DRL) in drone control and coordination~\cite{caballero2024artificial}, while specific DRL-based frameworks have implemented distributed policy learning from local observations and neighbour messages to support cooperative drone-swarm trajectory generation~\cite{Westheider2023MultiUAVDRL}. Iqbal \textit{et al.}~\cite{Iqbal2025} present a model that incorporates a Convolutional Neural Network (CNN) to estimate swarm coordination rates as indicators of trust, with each iteration using these estimates to classify drones as either trusted or malicious within the swarm. In visual drone navigation, supervised learning techniques have also been employed, combining optical flow methods with classifiers such as $k$-nearest neighbours (KNN) and support vector machines (SVM)~\cite{gong2023flight}.

For more than one decade, swarm communication among drones has been supported by the Internet of Drones (IoD) paradigm~\cite{internet/Hall16,adhoc/BoccadoroSG21,access/LabibBDB21}. Trust is a sine qua non in the IoD. Nair \textit{et al.}~\cite{vcomm/NairTP25} propose SoCoMNNet, which combines lightweight onboard Memristive Neural Networks with a trust-based SocioCognitive fuzzy system at the ground control station to detect GPS spoofing and distinguish it from normal mission deviations. To improve trust in IoD networks through intrusion detection, prior work has proposed a Q-learning-based, two-layer cooperative detection framework that uses dynamic voting across multiple nodes to strengthen host-level attack identification while reducing both false positives and false negatives~\cite{wu2023q}. Rathee \textit{et al.}~\cite{Rathee2022} introduce a trust-based secure communication framework for ad hoc UAV networks in smart-city settings, in which malicious devices are detected and excluded by evaluating historical interactions and assigning behavior-based local trust values recorded on a blockchain ledger.

Recent survey literature has also mapped the broader security landscape of UAV systems. Cordill \textit{et al.}~\cite{CordillFX25} review vulnerabilities and countermeasures across hardware, software, and communication layers, including privacy risks such as identity exposure and behavioural profiling. Similarly, Yu \textit{et al.}~\cite{YuKHE25} provide a taxonomy of UAV cyberattacks and countermeasures, covering threats such as spoofing, jamming, and coordination-level attacks, as well as mitigations including encryption, anomaly detection, and fail-safe mechanisms. At the communication layer, Wang \textit{et al.}~\cite{WangYMZXN25} propose a distributed segment-based routing framework with dynamic \textsc{SRv6} activation for scalable UAV swarm networking. 

While these works significantly advance cyber-defence and networking for UAV systems, they do not address the joint integration of formal mission-level coordination, explicit human authorisation, and hardware-enforced actuation safety under adversarial conditions.

\begin{figure*}[ht]
\centering
\includegraphics[width=\textwidth]{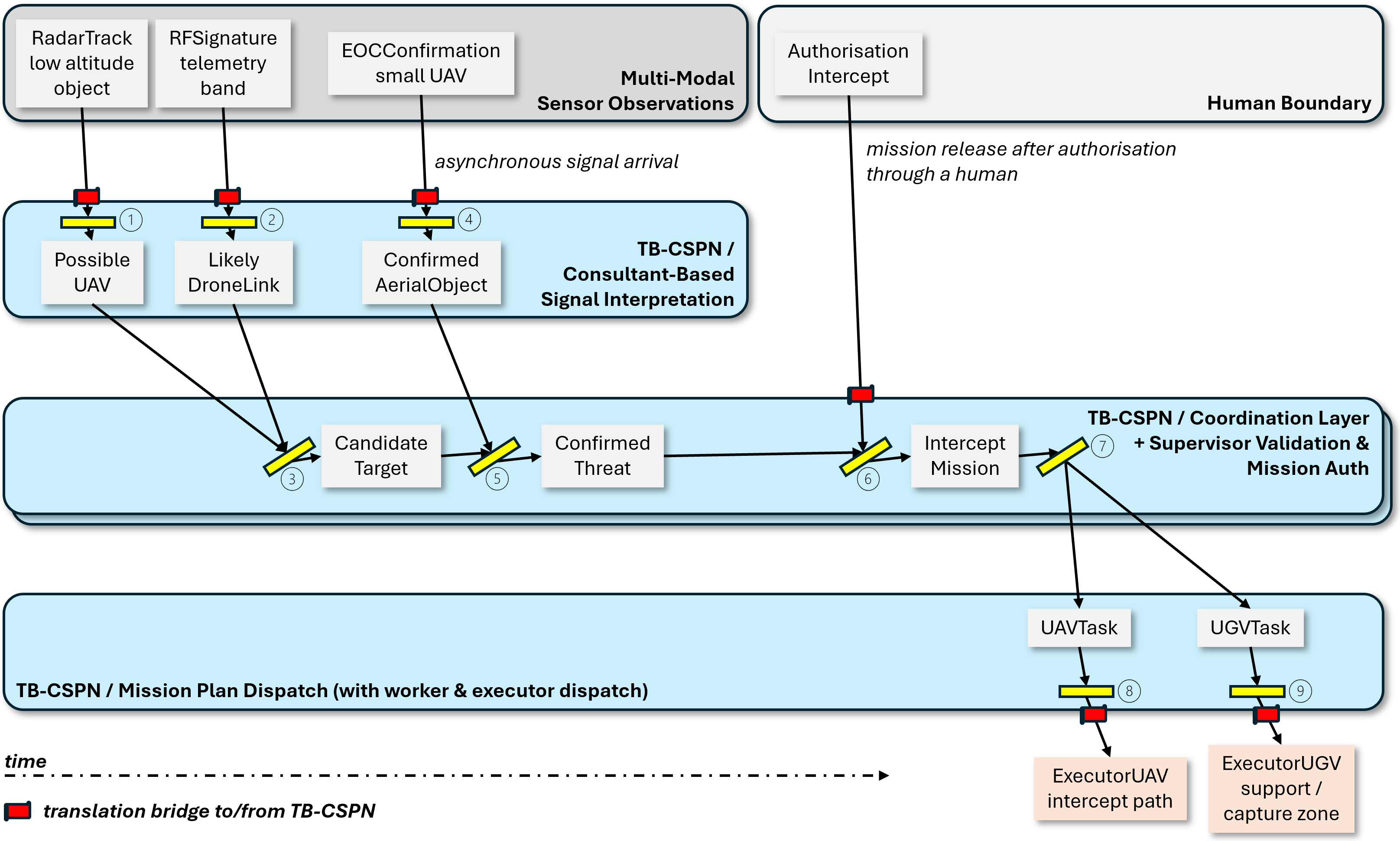}
\caption{Incremental mission formation in the coastal-surveillance case study. Heterogeneous sensor observations arrive asynchronously and are interpreted by consultant agents as semantic tokens, where labels such as \textit{PossibleUAV}, \textit{LikelyDroneLink}, and \textit{ConfirmedAerialObject} denote typed interpretations associated with specific topics.
TB-CSPN transitions perform both validation (through guards on signal consistency and timeliness) and semantic interpretation, progressively producing and refining semantic tokens, from \textit{CandidateTarget}  to \textit{ConfirmedThreat}. These tokens represent emergent semantic constructs arising from the integration of heterogeneous signals.
Once the required authorisation token is provided by the human-in-the-loop, the mission-release transition fires and produces an \textit{InterceptMission} token. A dispatch transition then generates task-specific tokens that worker agents consume, which are translated into platform-specific execution commands for the UAV and UGV executors.}
\label{fig:timeline}
\end{figure*}

\section{Case Study}
\label{sec:case_study}
To illustrate the proposed architecture, we consider a coastal surveillance scenario in which a heterogeneous swarm of UAV and UGV platforms monitors a maritime approach corridor for potential aerial threats. The environment is characterised by mixed civilian and operational traffic, intermittent communication conditions, and heterogeneous sensor signals.

The objective of the system is to detect, classify, and intercept potentially hostile aerial targets while maintaining safe actuation under uncertain or incomplete information \cite{WaltonKGCT22}; for more elaborate attack–defense confrontation scenarios, see~\cite{Chipade2023,Choi2022,LaarniVKLS22,LiuZZZ23}.

Figure~\ref{fig:timeline} illustrates a scenario in which a threat is detected by combining information from asynchronous sensor observations leading to mission formation and task dispatch. 
In this scenario, each transition in the depicted net is fired, as the two asynchronous signals detected by the altitude and telemetry sensors are interpreted by the logics embedded in the Transitions \textcircled{\small\textsf{1}} and \textcircled{\small\textsf{2}} as suspicious. 
Transition \textcircled{\small\textsf{3}} recognises their arrival within the established time span, and jointly interprets them as indicating the presence of a candidate target.
Combining this token with the one produced by Transition \textcircled{\small\textsf{4}} which interprets the electro-optical sensor output (also arriving within the relevant time window) as confirmation of the presence of a small UAV, Transition \textcircled{\small\textsf{5}}  produces a token indicating the presence of a threat.
As Transition \textcircled{\small\textsf{6}} receives this token as well as the authorisation to intercept issued by a human operator monitoring the process, it computes the plan for the intercepting mission, which Transition \textcircled{\small\textsf{7}} will map to specific instructions for the worker agents controlling the executors. 
Finally, the latter will activate the respective Transitions, \textcircled{\small\textsf{8}} and \textcircled{\small\textsf{9}}, to translate such instructions into executable commands.

Note that at any moment, lack of synchronisation or different interpretations of the situation would give rise to different scenarios, not leading to execution of intercepting actions.
Also note that tokens in the TB-CSPN layer carry typed semantic labels associated with domain-specific topics, rather than representing raw sensor data. Transitions, therefore, combine validation (e.g., temporal consistency and signal admissibility) with interpretation, and may or may not produce output tokens depending on guard conditions.
In this sense, tokens such as \textit{CandidateTarget} and \textit{ConfirmedThreat} do not correspond to directly observed entities but to emergent semantic constructs composed from multiple heterogeneous evidence sources.

\subsection{Coastal Surveillance Scenario}
The monitored area consists of a coastal perimeter equipped with multiple sensing systems. 
Multi-sensor low-altitude UAV surveillance commonly combines radar, electro-optical, acoustic, and radio sensing modalities to compensate for the limitations of any individual detector~\cite{Huang2022MultiSensorUA}. 
In this scenario, these modalities are realised through a coastal radar capable of detecting low-altitude aerial tracks, passive RF monitoring equipment for identifying potential drone communication signals, and a network of acoustic sensors deployed on buoys near the shoreline, while UAV platforms equipped with electro-optical (EO) cameras provide additional visual confirmation when required.

The swarm consists of two UAV platforms responsible for aerial interception and one UGV platform positioned near the shoreline for ground-based observation and potential target recovery. A coordination node hosts the TB-CSPN orchestration layer and supervisor agents.

Sensor systems generate asynchronous observations, which are propagated into the coordination architecture as tokens.

Sensor observations have the following structure:
\[
SensorObservations = \langle id, signal, t, source, confidence, ttl \rangle
\]

Typical signals in the scenario include
radar tracks indicating a low-altitude moving object,
RF emissions suggesting possible drone telemetry signals,
acoustic detections associated with propeller noise,
EO visual confirmations of a small aerial object.

Each observation contributes partial evidence regarding the identity of the detected object.

\subsection{Signal Interpretations}
Consultant agents interpret incoming observations through translation bridges and generate interpretation tokens representing candidate explanations.
For example:
\[
SignalInterpretation = \langle label, origin, t, confidence, ttl \rangle
\]

Possible interpretations can be labelled as
\textit{PossibleUAV},
\textit{LikelyDroneLink},
\textit{ConfirmedAerialObject}.

As additional sensor observations arrive, the coordination space progressively refines the set of active interpretation tokens. 
Radar tracks may initially elicit ambiguous interpretations, which are then strengthened or discarded when RF or visual confirmations become available.

When the required combination of evidence tokens and the authorisation token is present, the corresponding transition fires and produces a mission token:
\[
InterceptMission = \langle mission\_id, type, target, justification, t, ttl \rangle
\]

Once an \textit{InterceptMission} token is generated, the corresponding mission directive is consumed by the relevant worker agents to initiate the coordinated swarm action. Each worker agent consumes a task-specific token, such as \textit{UAVTask} or \textit{UGVTask}, and serves as a translation bridge across the digital--physical boundary, converting these coordination-level directives into vehicle-specific execution commands for the associated executor platform.

For example, a worker agent assigned to a UAV may convert the mission directive into trajectory waypoints or interception manoeuvres, while a worker agent associated with a UGV may generate navigation or observation commands. In this architecture, executor platforms themselves do not interpret mission tokens or perform mission-level reasoning; they simply execute the control instructions produced by their worker agents.

This separation preserves a clear distinction between coordination logic and physical execution. Mission interpretation and planning remain within the TB-CSPN coordination layer and its associated worker agents, while the executor platforms focus exclusively on deterministic, platform-level control tasks.

\subsection{Counterfactual Trajectories}
The trajectory described above traces one of several possible evolutions of the coordination space. The architecture's value lies equally in the trajectories that do \emph{not} lead to mission release. We illustrate three such cases.

\textit{Stale electro-optical confirmation.}
Suppose the EO sensor detects the suspected aerial object, but its observation token arrives after the temporal window established by Transition~\textcircled{\small\textsf{3}} has closed. The radar and RF tokens have expired or have been consumed by other transitions. The token produced by Transition~\textcircled{\small\textsf{4}} cannot be combined with a valid \textit{CandidateTarget} token, and Transition~\textcircled{\small\textsf{5}} does not fire. No \textit{ConfirmedThreat} token is generated. The architecture has, by structural means, prevented mission formation from incoherent evidence, without requiring a centralised arbiter to detect the inconsistency.

\noindent\textit{Authorisation withheld.}
Suppose evidence accumulates as before and a \textit{ConfirmedThreat} token is correctly produced. However, the supervisor agent declines to issue an authorisation token. This may be because the rules of engagement do not permit interception in the current operational context, or because the supervisor deems the threat assessment to be insufficient. Transition~\textcircled{\small\textsf{6}} cannot fire. The threat assessment persists in the coordination space, available for inspection and possible later authorisation, but no mission is dispatched. The architecture preserves the consultants' interpretive work while ensuring that no consequential action proceeds without explicit governance.

\textit{Conflicting interpretations.}
Suppose the radar consultant produces a \textit{PossibleUAV} token while the RF consultant interprets the same time window as containing only legitimate civilian telemetry, producing a \textit{LikelyCivilianTraffic} token. Both tokens enter the coordination space. The guard on Transition~\textcircled{\small\textsf{3}} requires concordant evidence; the conflicting interpretations are not consumed and the candidate-target transition does not fire. The conflict is preserved and visible in the net's marking, allowing supervisors or downstream consultants to inspect the disagreement and either await further evidence or invoke a dedicated conflict-resolution transition.
These three cases illustrate that the coordination architecture's correctness properties operate as much in the negative as in the positive: the structural guarantee is not merely that authorised missions follow valid evidence, but that unauthorised or incoherent missions \emph{cannot} be formed. This is the practical content of constrained non-determinism in the sensor-fusion setting.

\section{Protocol Integration in Practice}\label{sec:discussion}
In a practical deployment, communication would typically be realised through a layered rather than uniform end-to-end protocol stack.
The following discussion is intentionally abstract and illustrative, using prominent protocol families as examples. Operational (military) deployments may instead rely on proprietary or vendor-specific alternatives while preserving the same translation steps into and out of the TB-CSPN layers. 

At the \textit{multi-modal sensor observation} boundary, radar subsystems would commonly export detections or tracks through surveillance-oriented exchange formats such as \textsc{ASTERIX} \cite{ez2025analyzing}, which is defined by Eurocontrol as a data format for the exchange of surveillance-related information (\url{https://www.eurocontrol.int/asterix}). 
Electro-optical subsystems in IP-based installations would commonly rely on interoperable video-device interfaces such as \textsc{ONVIF-Profile~S} (Open Network Video Interface Forum) (\url{https://www.onvif.org/profiles/profile-s/}). 
Lin \textit{et al.}~\cite{LinCSCY18} describe a robust extension for seamless surveillance.
Passive RF and acoustic subsystems would more typically employ device-specific interfaces and signalling formats. Within the present architecture, however, these protocol-level representations are not directly consumed by the TB-CSPN layers. 
Instead, they are first processed by protocol adapters (or \textit{translation bridges} as shown in Figure~\ref{fig:timeline}) that normalise heterogeneous sensor outputs into internal observation messages or topics carrying the attributes required by coordination, including source, timestamp, confidence, and signal type. 
Consultant agents then operate on these normalised representations and translate them into semantic tokens, such as \textit{PossibleUAV}, \textit{LikelyDroneLink}, or \textit{ConfirmedAerialObject}. 
These semantic tokens, rather than the raw sensor messages themselves, constitute the actual inputs to the TB-CSPN layer and define the enabling conditions under which transitions may fire, once the relevant guards on timing, admissibility, and consistency are satisfied. 
This mechanism induces a coordination regime with constrained non-determinism: multiple transitions may be enabled, but the set of admissible evolutions remains strictly bounded by token availability, guard conditions, and temporal constraints. 
From an engineering perspective, it is useful to distinguish this form of non-determinism from both \emph{don't-care non-determinism}, where alternative execution orders lead to equivalent outcomes, and from unconstrained generative non-determinism, as exhibited by large language models, where the space of possible responses is open-ended. 
The design objective of TB-CSPN is to retain structured flexibility in coordination while avoiding both irrelevant variability and uncontrolled divergence in system behaviour.

A technically plausible middleware for this integration layer is a publish/subscribe architecture based on \textsc{ROS~2} with \textsc{DDS}-style message exchange, since such middleware supports the structured transport of heterogeneous observations while preserving the separation between protocol handling and TB-CSPN coordination logic. In the \textsc{PX4} ecosystem, for example, communication between \textsc{ROS~2} and the autopilot uses middleware that implements the \textsc{XRCE-DDS} protocol (eXtremely Resource Constrained Environments-Data Distribution Service), exposing \textsc{PX4} \textsc{uORB} messages as \textsc{ROS~2} messages and types and thereby allowing \textsc{ROS~2} workflows to access vehicle state information and issue commands (\url{https://docs.px4.io/main/en/ros2/}). 

The coordination node, therefore, functions not merely as a communication hub but as a protocol-convergence and semantic-lifting layer in which raw detections are progressively transformed into coordination-relevant abstractions, including in swarm settings~\cite{LeeKIISE2025}. 
This role is particularly important in \textsc{PX4}--\textsc{ROS~2} configurations based on \textsc{uXRCE-DDS} (\url{https://docs.px4.io/main/en/middleware/uxrce_dds}), where a client on \textsc{PX4} and an agent on the companion computer exchange data over serial or UDP links, publish selected \textsc{uORB} topics into the global \textsc{DDS} space, and thereby create the conditions under which communication overhead and scalability constraints become more pronounced as the number of participating nodes increases.

A corresponding translation principle applies at the \textit{digital--physical} boundary, where worker agents and executors implement the dispatch process, and physical actuation occurs.
Communication with aerial executors would most naturally use \textsc{MAVLink} (Micro Air Vehicle Link)~\cite{KoubaaAAJBK19}, which the official developer documentation describes as a lightweight messaging protocol for communication with drones and among onboard components (\url{https://mavlink.io/en/}). 
Yet, in the present architecture, the TB-CSPN layers do not emit protocol-level actuator messages directly. Rather, when mission-formation and dispatch transitions fire, they produce mission and task tokens that are consumed by worker agents, which are responsible for translating coordination-level outcomes into platform-specific control instructions, such as waypoint assignments, interception trajectories, loiter commands, navigation goals, or observation tasks, and only thereafter encoding these instructions in the protocol required by the corresponding executor platform. 

Where \textsc{PX4}-based vehicles are employed, this translation may proceed either through direct \textsc{MAVLink} command channels or through the \textsc{uXRCE-DDS} bridge, which \textsc{PX4} documents as exposing onboard messages to companion-computer workflows as though they were \textsc{ROS~2} topics. 
For experiments focusing on latency, throughput and packet loss, see~\cite{solpan2023dds}.

UGV platforms are typically more heterogeneous: a \textsc{ROS~2}/\textsc{DDS} interface is realistic for research platforms and service robots, whereas autopilot-style ground vehicles may also rely on \textsc{MAVLink} or vendor-specific fieldbus or Ethernet protocols. 
Accordingly, the communication model implied by the present architecture is best understood not as a single protocol chain but as a sequence of semantic transformations: from sensor-specific protocols to internal observation topics, from observation topics to semantic tokens enabling TB-CSPN transitions, and from mission or dispatch tokens to executor-specific command messages for UAV and UGV platforms.

\section{Discussion}\label{sec:discussion-proper}
The architecture proposed in this paper rests on three commitments that we wish to make explicit, since each invites legitimate questions and represents a deliberate trade-off rather than a settled answer.

\textit{The role of formal coordination in agentic AI.}
A reasonable alternative to TB-CSPN-based coordination would be to rely on direct LLM-mediated agent interaction, as exemplified by frameworks such as LangGraph, AutoGen, or ReAct-style loops. These approaches achieve flexibility through unconstrained natural-language exchange between agents. We do not dispute their utility for rapid prototyping or open-ended task execution. We argue, however, that they are unsuited to the present setting, where evidence must be fused under temporal constraints, where decisions must be auditable after the fact, and where governance must be structural rather than emergent. The cost of formal coordination is reduced flexibility; the benefit is that the conditions under which the system fires a mission token are explicit, inspectable, and bounded. In contested environments, we judge this trade-off as forced: an unauditable interception decision is not an acceptable engineering outcome, regardless of how naturally it was generated.

\textit{The locus of human authority.}
The architecture treats human authorisation as a token within the coordination layer rather than as an external veto over an otherwise autonomous loop. This choice has two consequences. First, it makes human authority \emph{structural}: a mission cannot fire without the appropriate token, and the absence of authorisation is itself a recorded state of the coordination space. Second, it commits the architecture to bounded-rate human involvement: the supervisor's authority is over the conditions of legitimate action, not over each individual transition. In settings where events outpace deliberation, this becomes critical. The supervisor authorises the envelope; the analogue safety layer enforces it; the digital coordination layer operates within it. We acknowledge that this distribution of responsibility differs from the simpler ``human-in-the-loop'' model often assumed in regulatory discussions, and that its alignment with frameworks such as the EU AI Act or NIST AI RMF requires further analysis we do not undertake here.

\textit{The boundaries of architectural reasoning.}
The contribution of this paper is architectural. 
The properties we claim---bounded non-determinism, temporal coherence of evidence, structural governance, hardware-enforced actuation safety---are properties of the architecture, not of any particular instantiation of it. 
A given deployment may fail to realise these properties through implementation flaws, sensor calibration errors, or supervisor misjudgement. What the architecture provides is not a guarantee that such failures will not occur, but a structure within which they can be localised, detected, and reasoned about. This is, we believe, the right bar for an architectural contribution; empirical demonstration that specific instantiations meet specific quantitative targets is a separate and necessary line of work, addressed in part by~\cite{FUTUREINTERNET/BorghoffBP26} and pursued further in the directions outlined in the Conclusion.
Within these commitments, the architecture offers a coherent answer to the question of how dependable autonomy can be achieved in cyber-physical settings where neither pure software autonomy nor pure human control is sufficient. The answer it offers is structural: separate interpretation from coordination from execution, govern the boundaries between them explicitly, and place irreducible safety constraints in a substrate that operates independently of the digital decision layer.

\section{Conclusion}\label{sec:conclusion}
This paper presents a sensor-driven coordination architecture for heterogeneous UAV/UGV swarms, in which uncertain, asynchronous and multimodal observations are transformed into mission-level actions under explicit governance and hardware-enforced safety constraints. The central architectural premise is that dependable swarm operation in contested environments necessitates a clear distinction between three frequently conflated concerns: the semantic interpretation of sensor data, coordination and mission formation, and physical execution at the actuator boundary. In the proposed approach, consultant agents interpret heterogeneous radar, radio frequency (RF), acoustic and electro-optical (EO) observations and convert them into semantic tokens. The TB-CSPN layer then governs the synchronisation and evolution of these tokens through guarded, time-bounded transitions. Finally, worker agents translate authorised mission tokens into control commands specific to the executor, while analogue safety envelopes independently enforce irreversible physical constraints.

The contribution of this paper is architectural. We do not present quantitative sensor-fusion benchmarks or hardware-in-the-loop validation at scale: those belong to a separate empirical programme. 
Here, we present the integration pattern that lends meaning to such empirical work: a coordination architecture in which sensor evidence enters with explicit semantics, evolves under formal constraints and is only actuated through governed mission tokens and hardware-enforced safety envelopes. The architectural separation of interpretation, coordination, and execution is what allows benchmarks of any one layer to be interpreted unambiguously.

The coastal surveillance case study demonstrates how the architecture can help shape missions gradually using distributed, partly uncertain evidence. In particular, it shows how asynchronous sensor observations can be normalised and interpreted semantically to create stronger coordination states, such as \textit{CandidateTarget}, \textit{ConfirmedThreat} and, ultimately, \textit{InterceptMission}. This progression clarifies the role of the TB-CSPN layer as an auditable semantic backbone linking sensing, governance, and execution. The study also shows how the digital and analogue layers complement each other: the former structures decision-making in situations of uncertainty, while the latter ensures that unsafe commands cannot reach the platform actuators in the event of software faults, cyber compromises or communication degradation.

Several limitations remain. The present study focuses on architecture and systems integration rather than on quantitative benchmarking, formal proof extensions, or large-scale hardware-in-the-loop validation. Furthermore, the communication mechanisms discussed in the paper are illustrative, and the protocol families used in real deployments may differ between vendors and mission environments. 

Future work will therefore follow three approaches: first, we will perform larger-scale simulations and comparative evaluations under conditions of degraded communication and adversarial interference; second, we will design and execute hardware-in-the-loop experiments using representative UAV/UGV platforms and analogue safety boards; and third, we will provide a deeper formal analysis of token lifetimes, coordination invariants and mission-level correctness conditions. 

Taken together, this will strengthen the empirical and formal basis of Guarded Swarms, in accordance with the main insight that dependable autonomy in contested cyber-physical environments is best achieved by combining semantically governed coordination with independently enforced physical safety.

\subsection*{Acknowledgments}
Figure~\ref{fig:conceptual} was generated by \textsc{Notebooklm} based on input from the authors, who reviewed and edited the output and take full responsibility for the content.
The authors wish to acknowledge the use of \textsc{DeepL Write} and \textsc{ChatGPT 5.2} in the writing of this paper. 
These tools were used to improve the language quality of the paper. 
Of course, it remains an accurate representation of the authors' underlying work and novel intellectual contributions.


\end{document}